\documentclass[11pt,a4paper]{article}
\usepackage[applemac]{inputenc}
\usepackage{watermark}
\usepackage{slashbox}
\usepackage[T1]{fontenc}
\usepackage[dvips]{graphicx}
\usepackage{psfrag}
\usepackage{pstricks}
\usepackage{acronym}
\usepackage{latexsym}
\usepackage[]{algorithm2e}
\usepackage[english]{babel}
\usepackage{epsfig,array,amsmath,amssymb,float}
\usepackage{xspace,calc,multirow,rotating,theorem}
 \usepackage{setspace}

\usepackage{longtable,calc,multirow}
\DeclareMathAlphabet{\mathbfeul}{U}{eur}{b}{n}
\DeclareMathAlphabet{\matheul}{U}{eur}{m}{n}
\DeclareMathAlphabet{\mathbfeus}{U}{eus}{b}{n}
\DeclareMathAlphabet{\matheus}{U}{eur}{m}{n}
\DeclareMathAlphabet{\mathbfzc}{OML}{pzc}{b}{n}
\DeclareMathAlphabet{\mathzc}{OT1}{pzc}{m}{it}
\DeclareMathAlphabet{\mathag}{OT1}{pag}{m}{n}
\DeclareMathAlphabet{\mathbfeuf}{U}{euf}{b}{n}
\DeclareMathAlphabet{\matheuf}{U}{euf}{m}{n}

\newcommand {\tr} {\rm\scriptscriptstyle T}

\newcommand{\beali}[1]{\begin{equation}\begin{aligned}#1\end{aligned}\end{equation}}

\newcommand{\bea}[1]{\begin{eqnarray} #1 \end{eqnarray}}

\newcommand{\bMat}[1]{\begin{bmatrix}#1\end{bmatrix}}

\newcommand {\eqrefn}{Eq.~\eqref}
\newcommand{\rank}[1]{\operatorname{rank}\left(#1\right)}  
  
  \newcommand{\fbf}{\mathbf{f}}
 \newcommand{\hbf}{\mathbf{h}} 
  \newcommand{\lbf}{\mathbf{l}}
  
  \newcommand{\rbf}{\mathbf{r}}
  
 \newcommand{\wbf}{\mathbf{w}} \newcommand{\xbf}{\mathbf{x}}

\newcommand{\Abf}{\mathbf{A}}  
\newcommand{\Dbf}{\mathbf{D}} \newcommand{\Ebf}{\mathbf{E}} 
\newcommand{\Gbf}{\mathbf{G}} \newcommand{\Hbf}{\mathbf{H}} \newcommand{\Ibf}{\mathbf{I}}
 \newcommand{\Kbf}{\mathbf{K}} 
  
\newcommand{\Pbf}{\mathbf{P}} \newcommand{\Qbf}{\mathbf{Q}} 
 \newcommand{\Tbf}{\mathbf{T}} 
 \newcommand{\Wbf}{\mathbf{W}} 
 \newcommand{\Zbf}{\mathbf{Z}} 

\newcommand{\zeros}{\boldsymbol{0}}\newcommand{\Rbb}{\mathbb{R}}\newcommand{\Sbb}{\mathbb{S}} \newcommand{\Zbb}{\mathbb{Z}}
{\theoremstyle{break}\theoremheaderfont{\normalfont\bfseries}}
{\theoremstyle{plain}\theorembodyfont{\normalfont\rmfamily}}
{\theoremstyle{break}\theoremheaderfont{\normalfont\bfseries}
{\theoremstyle{break}\theoremheaderfont{\normalfont\bfseries}
{\theoremstyle{break}\theoremheaderfont{\normalfont\bfseries}
{\theoremstyle{break}\theoremheaderfont{\normalfont\bfseries}
{\theoremstyle{break}\theoremheaderfont{\normalfont\bfseries}

{\theoremstyle{break}\theorembodyfont{\normalfont\rmfamily}}
{\theoremstyle{break}\theorembodyfont{\normalfont\rmfamily}}

\newcommand {\vlightrule}{\kern1ex\vrule width0.2pt\kern1ex}
\definecolor{GreenByNirav}{rgb}{0.2,0.5,0.21}

\newcommand{\vectorize}[1]{\operatorname{vec}(#1)}  

\def\endkeywords{\vspace{0.6em}\par\if@twocolumn\else\endquotation\fi
    \normalsize\rm}

\begin{document}
\title{Monotonous (Semi-)Nonnegative Matrix Factorization}
\author{
        \begin{tabular}[t]{cccc}
     Nirav Bhatt$^{a\star}$, & Arun Ayyar$^{b}$\footnote{Corresponding authors. E-mail: {\tt niravbhatt@iitm.ac.in,\, arun.ayyar@gmail.com}}
        \end{tabular}
	\\
	$^a$Systems $\&$ Control Group, Department of Chemical Engineering, \\ Indian Institute of Technology Madras, Chennai - 600036, India\\
	$^b$ Santa Fe Research, IIT Madras Research Park, Chennai - 600036, India
}
\maketitle

\date{}

\begin{abstract}

Nonnegative matrix factorization (NMF) factorizes a  non-negative  matrix  into  product of two non-negative matrices, namely a signal matrix and a mixing matrix. NMF suffers from the scale and ordering ambiguities.  Often, the source signals can be monotonous in nature. For example, in source separation problem, the source signals can be monotonously increasing or decreasing while the mixing matrix can have  nonnegative entries.  NMF methods may not be effective for such cases as it suffers from the ordering ambiguity.  This paper proposes an approach to incorporate  notion of monotonicity in NMF, labeled as { \it monotonous NMF}. An algorithm based on alternating least-squares is proposed for recovering monotonous signals from a data matrix.  Further, the assumption on mixing matrix is relaxed to extend monotonous NMF for  data matrix with real numbers as entries. The approach is illustrated using synthetic  noisy data.  The results obtained by monotonous NMF are compared with standard NMF algorithms in the literature, and it is shown that monotonous NMF estimates source signals well in comparison to standard NMF algorithms when the underlying sources signals are monotonous.
\end{abstract}

\section{Introduction}
Nonnegative matrix factorization (NMF) is one of the widely used matrix factorization techniques with application areas ranging from basic sciences such as  chemistry, environmental science, systems biology to   image and video processing, blind source separation, text mining, social network analysis \cite{Cichocki09,DingTJ10,PaateroT94,Virtanen07}.  The reasons of wide applicability of NMF are two folds: (i) Non-negativity feature is prevalent in real world data, and (ii) the latent factors estimated by NMF are easily interpretable. Further, the seminal paper   by Lee and Seung \cite{LeeS99,LeeH01} has helped in popularizing NMF in various fields.

In unsupervised learning methods, the objective is to extract  signals or   features from the given data. For example, in blind source separation (BSS), the objective is to identify the underlying source signals from noisy data.
 NMF has been routinely applied to separate source signals from  noisy data  \cite{Cichocki09,RapinBLS13}. NMF decomposes a data matrix into a mixing matrix (coefficients of signals), and a source signal matrix.  Since NMF suffers from ordering and scaling ambiguities, NMF factorization is not unique \cite{Cichocki09,RapinBLS13}.  To obtain appropriate  solution,  constraint such as sparsity has been incorporated in NMF \cite{Cichocki09,Hoyer04}. Further, semi-NMF imposes non-negativity constraints only on the source signals and allows the mixed sign entries in  data and mixing matrices \cite{DingTJ10}.   
 
 Many algorithms have been proposed to improve the speed and convergence for   NMF.  One of the most commonly used algorithms is multiplicative update rules and its variants proposed in \cite{BerryBLPP07,LeeH01}. Recently, several algorithms in alternating least-squares framework combined with numerical optimization techniques, active-set method \cite{KimP11},  projected gradient methods \cite{Lin07}, quadratic programming \cite{CichockiZ08} etc., have been proposed to improve speed and convergence  for NMF. Further, necessary and sufficient conditions for unique NMF decomposition have been also investigated \cite{DonohoS03,HuangSS14}.  Although there have been considerable work on improvement of algorithms  for better solutions, best of our knowledge, there is  no attempt on solving monotonicity in signals in NMF. 
 
 Source signals in chemistry and systems biology often exhibit monotonous   property. In such scenarios, the formulation of monotonicity constraints is important for recovering source signal using NMF or semi-NMF.  In this paper, it is proposed to investigate how to resolve monotonicity in NMF. We will propose a new approach, called monotonous NMF, for recovering monotonous source signals from noisy data. Further, we will extend this approach to semi-NMF.  Two algorithms for monotonous (semi-) 
 NMF are also proposed. The approach is demonstrated on synthetic data. Moreover, future work on  incorporation of sparseness and integer entries in mixing matrix will be discussed.

 The rest of this paper is organized as follows. In Section~\ref{Sec:MF}, the matrix factorization and NMF related background are introduced. We propose  extensions of NMF which resolves monotonicity of source signal in Section~\ref{Sec:MNMF}. Section~\ref{Sec:Example} illustrates  monotonous NMF on simulation studies based on synthetic data sets. Further, it compares the performance of the proposed monotonous NMF with two NMF algorithms in the existing literature.  Section~\ref{Sec:Conclusion}  concludes the paper.
 Next, the notations used in the paper are described.

\subsection{Notations}
The bold capital and small letters are used for defining  matrices and vectors; the small italic letters are used to define scaler quantities; $\wbf_{.,i}$ and $\hbf_{i,.}$ are used for denoting column and row vectors, respectively; $\Qbf(i,j)$ denotes $(i,j)th$ element of matrix $\Qbf$; $p(.)$ indicates permutations;  $\vectorize{\Qbf}$ indicates the vectorization of a matrix $\Qbf$ which converts the matrix into a column vector; $(\cdot)^{+}$ indicates pseudo-inverse of matrix; $\Qbf^{\tr}$ is  transpose of matrix $\Qbf$; $\Rbb^{m \times n}$ indicates  $m \times n$-dimensional matrices; $\|\cdot\|$ denotes Frobenius norm; $\|\cdot\|_1$ denotes $L_1$-norm of a vector;  $\Zbb$ is integers; $\Rbb_{+}$ indicates nonnegative real numbers; $\Sbb_{+}^n$ denotes $n \times n$-dimensional symmetric positive semidefinite matrices; $\otimes$ denotes Kronecker products; $\Ibf_{n}$ is an $n$-dimensional identity matrix; $\zeros_{n}$ denotes vector of length $n$ with zeros as elements;  $\geq or \leq$ indicates component-wise inequality for matrices and vectors.

\section{Matrix Factorization}\label{Sec:MF}

 A data matrix $\Zbf \in \Rbb^{n \times m}$ can be factorized as product of $\Wbf \in \Rbb^{n \times s}$ and $\Hbf \in \Rbb^{s \times m}$ for $s<< \min(m,n)$ as follows:
\bea{\Zbf\approx \Wbf\,\Hbf. \label{Eq:MatFac} }
In BSS problems,  $\Wbf$ is unknown mixing matrix and $\Hbf$ is source signal matrix containing unknown signals. Then, $m$ is the number of observations, $n$  the number of samples, and $s$ is the number of sources. Note that \eqrefn{Eq:MatFac} considers approximate factorization of the data matrix. Since physical systems are often corrupted by noise, the exact factorization of $\Zbf$  into $\Wbf$ and $\Hbf$  of the data should not be attempted. In presence of noise, the exact factorization of $\Zbf$ can be written as
\bea{\Zbf = \Wbf\,\Hbf +\Ebf, \label{Eq:ExactMatFac} }
where $\Ebf$ is an $m \times n-$dimensional matrix containing a contribution of noise in a given data. 

If $\Wbf$ is known and $\rank{\Wbf}=s$, then the least-squares solution of the unknown signal matrix is given by $\hat\Hbf=\Wbf^{+}\Zbf$. However, $\Wbf$ is often not known, and the objective of BSS problems is to estimate the unknown mixing matrix and signal matrix from the given data matrix.  Note that BSS problems can  be also viewed as unsupervised learning methods where  the objective is to extract useful features from the given data. To accomplish this task,  a prior knowledge about the underlying system  such as non-negativity, and sparsity are imposed as constraints. These constraints lead to factorization of the data matrix with  meaningful physical insights. The non-negativity constraint leads to popular non-negative  matrix factorization (NMF).  In the next section, NMF is described briefly. 

\subsection{Nonnegative matrix factorization (NMF)} 
 NMF aims to factorize the $\Zbf$ into two nonnegative matrices $\Wbf \in \Rbb^{n \times s}_{+}$ and $\Hbf \in \Rbb^{s \times m}_{+}$. In NMF, the following problem is solved to obtain $\Wbf$ and $\Hbf$
\beali{& \min_{\Wbf,\,\Hbf}\,\|\Zbf - \Wbf\Hbf \| \\
&\text{st}\,\, \Wbf \geq \zeros,\,\, \Hbf \geq \zeros. \label{Eq:NMFFact}} 
The imposing  of nonnegative constraints leads to only additive combinations of several rank-one factors to represent the data. Several algorithms have been proposed in the literature to solve \eqrefn{Eq:NMFFact}, efficiently. However, NMF solution suffers from the following ambiguities \cite{RapinBLS13}:
\begin{itemize}
\item Scale ambiguity: Consider a non-singular matrix $\Tbf \in \Rbb^{s \times s}_{+}$. Then, the factorization of $\Zbf$ can also represented as:
 \beali{\Zbf\approx \Wbf\Hbf =(\Wbf\Tbf)(\Tbf^{-1}\Hbf)=\Wbf_{T}\Hbf_{T}. \label{Eq:NMFFactScaleAmb}}
\eqrefn{Eq:NMFFactScaleAmb} indicates that $\Wbf_{T}$ and $\Hbf_{T}$ are also solution of the problem \eqref{Eq:NMFFact}. Hence, NMF leads to a local optimum of the solution depending on the initialization, and provides numerous solutions to  the problem \eqref{Eq:NMFFact}. 
\item Ordering: Note that $\Zbf \approx \sum_{i} \wbf_{.,i}\hbf_{i,.}=\sum_{i} \wbf_{.,p(i)}\hbf_{p(i),.}$ with $p(\cdot)$ is an ordering (or permutation). Hence, it is not possible to recover the order of the columns of $\Wbf$ and of the rows $\Hbf$.
\end{itemize}
 Therefore, the formulation of NMF in \eqrefn{Eq:NMFFact} has many solutions.  Additional constraints such as sparsity, partial or full knowledge of noisy $\Wbf$ need to be imposed to reduce the number solutions or to obtain a unique solution of the problem \eqref{Eq:NMFFact}. These constraints depend on the applications and underlying physical system. Often, the signal sources are    in increasing or decreasing orders , i.e., monotonous in nature. In such case, each row of $\Hbf$ contains ordered elements. Hence, the monotonous constraint  can be added in the optimization problem \eqref{Eq:NMFFact}   for recovering these signals from the data matrix. The main objective of this paper is to develop an approach based on NMF for recovering monotonous  signal sources.  This  leads to a novel method, called monotonous NMF, which is described next.

\section{Monotonous NMF}\label{Sec:MNMF}
In this section, the monotonous constraints on the source signals will be formulated  and imposed on the factorization problem  \eqref{Eq:NMFFact}. Consider the $i${th} row of $\Hbf$ matrix, $\hbf_{i,.}$ with $i=1,\ldots, s$, whose elements are monotonously increasing or decreasing order.  Without loss of generality, we assume that out of the $s$ rows, the first $c$ rows have entries which are monotonously increasing, while the remaining rows have entries that are monotonously decreasing. Then, the following relationships hold between the elements:
\beali{h_{k,1}\leq h_{k,2}\leq\ldots\leq h_{k,m}, \, k=1,\ldots,c,\,\, \text{(increasing order)}\\h_{l,1}\geq h_{l,2}\geq\ldots\geq h_{l,m},\, l=c+1,\ldots,s,\,\, \text{(decreasing order)}\label{Eq:constraints}.}
The optimization problem in \eqrefn{Eq:NMFFact} can be reformulated by imposing the constraints  in \eqrefn{Eq:constraints} as: 
\beali{& \min_{\Wbf,\,\Hbf}\,\|\Zbf - \Wbf\Hbf \| \\
&\text{st}\,\, \Wbf \geq \zeros,\,\, \Hbf \geq \zeros,\\
& h_{k,1}\leq h_{k,2}\leq\ldots\leq h_{k,m}, \, k=1,\ldots,c,\\
& h_{l,1}\geq h_{l,2}\geq\ldots\geq h_{l,m}.\, l= (c+1),\ldots,s.\label{Eq:NMFMonoton}}
The  formulation in \eqrefn{Eq:NMFMonoton} is nonconvex optimization problem. The solution to the optimization problem \eqref{Eq:NMFMonoton} can be obtained by decomposing it into two subproblems.  These problems can be solved  alternately till  the solution converges. This approach is known as alternating least squares (ALS) in the literature \cite{KimP11,Lin07}.    The two subproblems  can be formulated as follows:

\begin{itemize}
\item Given $\Hbf$:
\beali{   \min_{\Wbf}\,\|\Zbf - \Wbf\Hbf \| \\
\text{st}\,\,  \Wbf \geq \zeros
\label{Eq:NMFMonotoninW}}
\item Given $\Wbf$:
\beali{ & \min_{\Hbf}\,\|\Zbf - \Wbf\Hbf \| \\
&\text{st}\,\,  \Hbf \geq \zeros\\
& h_{k,1}\leq h_{k,2}\leq\ldots\leq h_{k,m}, \, k=1,\ldots,c,\\
& h_{l,1}\geq h_{l,2}\geq\,\ldots\geq h_{l,m}.\, l=(c+1),\ldots,d. \label{Eq:NMFMonotoninH}}
\end{itemize}

Note that objective functions in Eqs.~\eqref{Eq:NMFMonotoninW} and \eqref{Eq:NMFMonotoninH} are quadratic functions. If we can formulate these subproblems in terms of standard quadratic programming as follows:
\beali{& \min_{\xbf}\frac{1}{2}\xbf^{\tr}\Pbf\xbf+ \fbf^{\tr}\xbf  \\
&\text{st}\,\,  \Gbf\xbf \leq \lbf \\
&\,\,\quad\Kbf\xbf=\rbf
 \label{Eq:QP}}
where $\xbf \in \Rbb^{p}$,  $\Pbf \in \Sbb_{+}^{p}$, $\Gbf\in \Rbb^{t \times p}$,  $\lbf \in \Rbb^{ p}$, $\Kbf\in \Rbb^{b \times p}$, and $\rbf  \in \Rbb^{b}$, then it can be shown that both subproblem are convex optimization problems \cite{BoydV04}. The standard ALS framework can be applied to solve the problems \eqref{Eq:NMFMonotoninW}--\eqref{Eq:NMFMonotoninH}. We reformulate  the problems \eqref{Eq:NMFMonotoninW}--\eqref{Eq:NMFMonotoninH} in terms of standard form \eqref{Eq:QP} as: \vspace{-0.2cm}
\begin{itemize}
\item Given $\Hbf$:
\beali{& \min_{\Wbf}\vectorize{\Wbf}^{\tr}(\Hbf\Hbf^{\tr} \otimes \Ibf_n) \vectorize{\Wbf}- 2\vectorize{\Zbf\Hbf^{\tr}}^{\tr}\vectorize{\Wbf} \\
&\text{st}\,\, \\ & -\Ibf_{ns}\vectorize{\Wbf} \leq \zeros_{ns} \label{Eq:quadfromW}}
\item Given $\Wbf$:
\beali{& \min_{\Hbf}\vectorize{\Hbf}^{\tr}(\Ibf_m \otimes \Wbf^{\tr}\Wbf) \vectorize{\Hbf}- 2\vectorize{\Zbf^{\tr}\Wbf}^{\tr}\vectorize{\Hbf} \\
&\text{st}\,\,\\&  \bMat{-\Ibf_{ms}\\\Abf}\vectorize{\Hbf} \leq \zeros_{ms} \label{Eq:quadfromH} }
\end{itemize}
 where $\Abf\in\Rbb^{(m-1)s \times ms}$ is matrix having the  form

$\Abf=\bMat{\Dbf_1&\\&\Dbf_2\\&&\ddots\\&&&\Dbf_s}$.  The entries not shown in $\Abf$ are zeros. The matrices $\Dbf_k\in\Rbb^{(m-1) \times m},\, k=1,2,\ldots
,c$ and $\Dbf_l\in\Rbb^{(m-1) \times m}, \, l=c+1,,\ldots, s$ are defined as follows: 
\beali{
\Dbf_k=\bMat{1 & -1 \\ & 1 & -1\\ & & \ddots&\ddots \\ &&& 1 &-1\\ &&&& 1 &-1}
\nonumber \\
\Dbf_l=\bMat{-1 & 1 \\ & -1 & 1\\ & & \ddots&\ddots \\ &&& -1 &1\\ &&&& -1 &1}}
Note that $\Dbf_k$ and $\Dbf_l$ are Toeplitz matrices with the first rows being $[1\,\,-1\,\,\, 0\, \ldots\, 0]$ and $[-1\,\, 1\,\, 0\, \ldots\, 0]$, respectively. 

Since $\Hbf \geq \zeros $ and $\Wbf \geq \zeros$, the matrices $(\Hbf\Hbf^{\tr} \otimes \Ibf_n)$ and   $(\Ibf_m \otimes \Wbf^{\tr}\Wbf)$ in  Eqs.~\eqref{Eq:quadfromW} and  \eqref{Eq:quadfromH} are positive semi-definite, and of full rank  $s$. Further, both subproblems~\eqref{Eq:quadfromW}--\eqref{Eq:quadfromH} are subjected to linear inequality  constraints. Hence, the objective problems in \eqref{Eq:quadfromW}--\eqref{Eq:quadfromH} are  quadratic programming (QP)  of form \eqref{Eq:QP}. Consequently, they are convex optimization problems \cite{BoydV04}. The optimal solutions of the problems in Eqs.~\eqref{Eq:quadfromW}--\eqref{Eq:quadfromH}  exist.   Standard QP  algorithms such as active-set methods, interior-point methods etc \cite{BoydV04,NocedalW99} can be used to solve the optimization problems \eqref{Eq:quadfromW}--\eqref{Eq:quadfromH} in the ALS framework. The  implementation of these algorithms is available in  in commercial scientific  packages.    
%

Next, we present an algorithm to solve  the optimization problems \eqref{Eq:quadfromW}--\eqref{Eq:quadfromH} for  estimating $\Wbf$ and $\Hbf$ given $\Zbf$. Further, one can impose the upper bounds on the elements of $\Wbf$ and $\Hbf$ based on the data matrix and the underlying physical problem. Algorithm \ref{Algo:MNMF} presents an implementation for solving monotonous NMF problem  \eqref{Eq:NMFMonoton}  in the ALS framework.
\begin{algorithm}[]
\caption{Algorithm for Monotonous NMF}\label{Algo:MNMF}
 \KwData{$\Zbf$, number of source signals ($s<< \min{(m,n)}$)}
 \KwResult{$\Wbf$, $\Hbf$}
 initialization: $\Wbf_{old}$, $\Hbf_{old}$ ;\\
 \While{$\|\Wbf_{old} -\Wbf_{new}\|\geq$ tolerance and $\|\Hbf_{old} -\Hbf_{new}\|\geq$ tolerance }{
  read current\;
  $\Wbf_{new} \Longleftarrow$ Solution of \eqref{Eq:quadfromW} \\
    $\Hbf_{new} \Longleftarrow$ Solution of \eqref{Eq:quadfromH} 
  }
  {\bf Return} $\Wbf =\Wbf_{new}$, $\Hbf=\Hbf_{new}$
 
\end{algorithm}

In Algorithm~\ref{Algo:MNMF},   $\Hbf$ can be normalized after each iteration. Then, $\Wbf$ has to be scaled in appropriate manner  before proceeding to next iteration. Further, note that the subproblem  \eqref{Eq:quadfromW} is a standard subproblem in NMF algorithms based on the ALS framework. Then, the multiplicative rules proposed in \cite{BerryBLPP07,LeeH01} can also be used  to update $\Wbf$ instead of solving the optimization \eqref{Eq:quadfromW} in the ALS framework. 
It has been shown that  the algorithms in the ALS framework  (such as Algorithm~\ref{Algo:MNMF}) decrease the function value at each iteration because the ALS framework produces stationary point at each iteration \cite{KimP11,Lin07}. Hence, Algorithm~\ref{Algo:MNMF} converges to a solution but note that the convergence speed may  be slower. 
\subsection{Monotonous Semi-NMF}
This section demonstrates how idea of monotonous NMF can be extended by relaxing nonnegative constraints on the mixing matrix, $\Wbf$ and $\Zbf$. This will allow to apply monotonous NMF to data matrix consisting negative entries. When we relax nonnegative constraints,  $\Wbf$ entries can have both positive or negative signs. 
 This kind of NMF is referred as {\em semi-NMF} in the literature \cite{DingTJ10}. However, the entries of signal matrix are still nonnegative and monotonous.  This relaxation on the mixing matrix allows the non-positive entries in data matrix. The optimization problem \eqref{Eq:NMFMonotoninW} can be reformulated without the non-negativity constraints as follows: 
\beali{   \min_{\Wbf}\,\|\Zbf - \Wbf\Hbf \|.
\label{Eq:NMFLSinW}}
For given $\Hbf$, \eqrefn{Eq:NMFLSinW} is a least-squares problem, and hence, the analytical update rule can be obtained as
\beali{\Wbf=\Zbf\Hbf^{\tr}(\Hbf\Hbf^{\tr})^{+}\label{Eq:UpdateRuleW}} 
In \eqrefn{Eq:UpdateRuleW}, $\Hbf\Hbf^{\tr}$ is  positive semidefinite matrix and its inverse (or pseudo-inverse) exists.  With this update rule, a modified version of Algorithm~\ref{Algo:MNMF} is given in Algorithm~\ref{Algo:MSemiNMF}.
\begin{algorithm}[]
\caption{Algorithm for Semi-NMF}\label{Algo:MSemiNMF}
 \KwData{$\Zbf$, number of source signals ($s<< \min{(m,n)}$)}
 \KwResult{$\Wbf$, $\Hbf$}
 initialization: $\Wbf_{old}$, $\Hbf_{old}$ ;\\
 \While{$\|\Wbf_{old} -\Wbf_{new}\|\geq$ tolerance and $\|\Hbf_{old} -\Hbf_{new}\|\geq$ tolerance }{
  read current\;
  $\Wbf_{new} \Longleftarrow$ $\Zbf\Hbf^{\tr}_{old}(\Hbf_{old}\Hbf^{\tr}_{old})^{+} $\\
    $\Hbf_{new} \Longleftarrow$ Solution of \eqref{Eq:quadfromH} 
  }
  {\bf Return} $\Wbf =\Wbf_{new}$, $\Hbf=\Hbf_{new}$
\end{algorithm}

This extension is in line of work done in \cite{DingTJ10}. Further,  Algorithm~\ref{Algo:MSemiNMF} converges to a solution as we have discussed for  Algorithm~\ref{Algo:MNMF}.  

\section{Illustrative example} \label{Sec:Example}
\begin{figure}
\begin{center}
\includegraphics[scale=0.40]{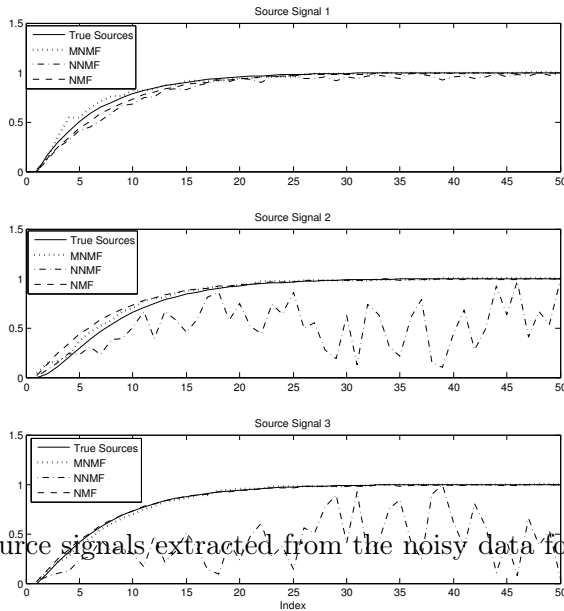}
\end{center}\vspace{-2cm}
\caption{Source signals extracted from the noisy data for Scenario S1} \label{Fig:MNMF}
\end{figure}

\begin{figure}
\begin{center}
\includegraphics[scale=0.40]{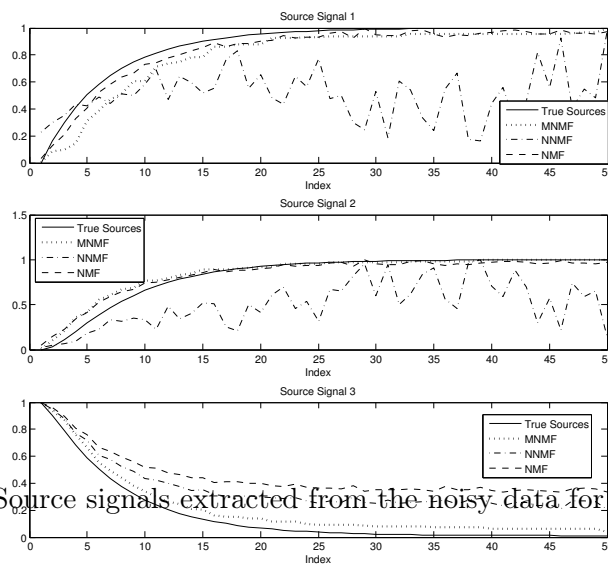}
\end{center}\vspace{-2cm}
\caption{Source signals extracted from the noisy data  for Scenario S2} \label{Fig:MNMFV2}
\end{figure}

In this section, we demonstrate monotonous NMF on  synthetic data involving three source signals. We have performed simulation studies for two scenarios: (S1)  monotonically increasing source signals, and (S2) mixed monotonous source signals (two increasing signals, one decreasing signal). The noise-free data matrix is constructed in the following manner. Three monotonically  source signals  are considered as shown in Figures~\ref{Fig:MNMF} and~\ref{Fig:MNMFV2} for Scenarios S1 and S2, respectively. Fifty sample points of signals are available in both scenarios. The mixing matrix  of size $(8 \times 3)$ is generated from  uniform distribution in the interval (0,1). The noise-free data matrix $\Zbf$ of size ($8 \times 50)$ is obtained by multiplication of  the mixing matrix and signal source matrix. To generate the noisy data, 5\% random uniformly distributed noise is added to the noisy-free data. All  simulations are performed in MATLAB. For Algorithm 1, "quadprog" function is used to solve the problems (10)-(11).

The following three methods are  applied to both scenarios: (i) Monotonous NMF (labeled MNMF), (ii) NMF using multiplicative rules (labeled NNMF) (MATLAB in-built function), and (iii) fast NMF (labeled NMF) algorithm based on active-set methods in \cite{KimP11} with the assumption of three source signals.

\paragraph{Scenario S1}
  The normalized  source signals obtained by applying  three methods along with the true one are shown in Figure~\ref{Fig:MNMF} for Scenario S1. The reconstruction errors ($\|\Zbf - \Wbf\Hbf \|$) are 
      0.1197,   0.1564 and,  0.9835 for MNMF, NNMF, and NMF, respectively. This result   shows that MNMF performs well in comparison of  NNMF and NMF. It should be noted that NMF estimates  rank-deficient source signal matrix ($\rank{\Hbf}=1$ for NMF). In other words, NMF factorizes the data matrix into single one-rank factor instead of  three one-rank factors, while MNMF and NNMF factorize the data matrix into three rank-one factors. However, NNMF fails to capture monotonous behaviour of source signals. 
  \paragraph{Scenario S2}
 The normalized  source signals obtained using the three methods along with the true one are shown in Figure~\ref{Fig:MNMFV2} for Scenario S2. The reconstruction errors ($\|\Zbf - \Wbf\Hbf \|$) are 
      0.1189,   0.1399 and,  0.5736 for MNMF, NNMF, and NMF, respectively. In this case, MNMF performs better in comparison to other methods to capture both kinds of monotonous signals. In this case, NMF estimates rank-deficient source signal matrix, however, the rank of $\Hbf$ has increased by one $\rank{\Hbf}=2$ for NMF. Note that NNMF fails to capture monotonous behaviour of source signals in Scenario 2 too. 

\section{Conclusions }\label{Sec:Conclusion}
The paper has proposed an approach to incorporate notion of monotonicity in NMF. The new extension is called as monotonous NMF. Monotonous NMF has been applied to recover  monotonous source signals from noisy data. Further, we have extended monotonous NMF to semi-NMF by relaxing non-negativity constraints on the mixing matrix. Algorithms for monotonous (semi-)NMF using quadratic programming have been proposed in the ALS framework. The illustrative examples show that the monotonous NMF performs better in comparison to the algorithms of NMF in the literature when the source signals exhibit monotonous behaviour. This indicates the importance of monotonicity constraint in NMF.

\bibliography{ArCornell}

\end{document}